\documentclass[10pt,twocolumn,letterpaper]{article}

\usepackage{iccv}
\usepackage{times}
\usepackage{epsfig}
\usepackage{graphicx}
\usepackage{amsmath}
\usepackage{amssymb}

\usepackage{subcaption}
\usepackage{booktabs}
\usepackage[ruled,vlined]{algorithm2e}
\usepackage{algorithmic}
\usepackage{multirow}
\usepackage{diagbox}
\usepackage{float}

\usepackage{authblk}
\makeatletter
\renewcommand\AB@affilsepx{\quad\protect\Affilfont}
\makeatother

\usepackage[pagebackref=true,breaklinks=true,letterpaper=true,colorlinks,bookmarks=false]{hyperref}

 \iccvfinalcopy 

\def\iccvPaperID{6887} 

\ificcvfinal\fi

\newcommand\x{0.23}

\newcommand\nnfootnote[1]{%
  \begin{NoHyper}
  \renewcommand\thefootnote{}\footnote{#1}%
  \addtocounter{footnote}{-1}%
  \end{NoHyper}
}

\begin{document}


\title{FullFormer: Generating Shapes Inside Shapes}

\author[1]{Tejaswini Medi$^*$}
\author[2]{Jawad Tayyub$^*$}
\author[3]{Muhammad Sarmad$^*$}
\author[3]{Frank Lindseth}
\author[1,4]{Margret Keuper}
\affil[1]{University of Siegen}
\affil[2]{Endress + Hauser}
\affil[3]{Norwegian University of Science and Technology}
\affil[4]{Max Planck Institute for Informatics and Saarland Informatics Campus}

\maketitle

\nnfootnote{*Equal Contribution}

\ificcvfinal\thispagestyle{empty}\fi
\begin{abstract}
Implicit generative models have been widely employed to model 3D data and have recently proven to be successful in encoding and generating high-quality 3D shapes. This work builds upon these models and alleviates current limitations by presenting the first implicit generative model that facilitates the generation of complex 3D shapes with rich internal geometric details. To achieve this, our model uses unsigned distance fields to represent nested 3D surfaces allowing learning from non-watertight mesh data. We propose a transformer-based autoregressive model for 3D shape generation that leverages context-rich tokens from vector quantized shape embeddings. The generated tokens are decoded into an unsigned distance field which is rendered into a novel 3D shape exhibiting a rich internal structure. We demonstrate that our model achieves state-of-the-art point cloud generation results on popular classes of 'Cars', 'Planes', and 'Chairs' of the ShapeNet dataset. Additionally, we curate a dataset that exclusively comprises shapes with realistic internal details from the `Cars' class of ShapeNet and demonstrate our method's efficacy in generating these shapes with internal geometry.
\end{abstract}

\section{Introduction}
\label{sec:intro}
Continuous representations of data and signals in the form of implicit functions are impacting many research areas of computer vision and graphics. The idea of having a continuously learned function to represent 3D data implicitly is efficient since these functions can represent diverse topologies as well as being agnostic to resolution \cite{chibane2020implicit}. Neural networks have been successfully utilized to parameterize these implicit functions. Applications of neural implicit representation are widespread \eg geometry representation \cite{occupancy,atzmon2020sal,DeepSDF},  image super-resolution \cite{chen2019learning} and generative models \cite{niemeyer2021giraffe,schwarz2020graf,zeng2022lion} etc.

\begin{figure}[t]
  \centering
  \includegraphics[keepaspectratio, width=\columnwidth]{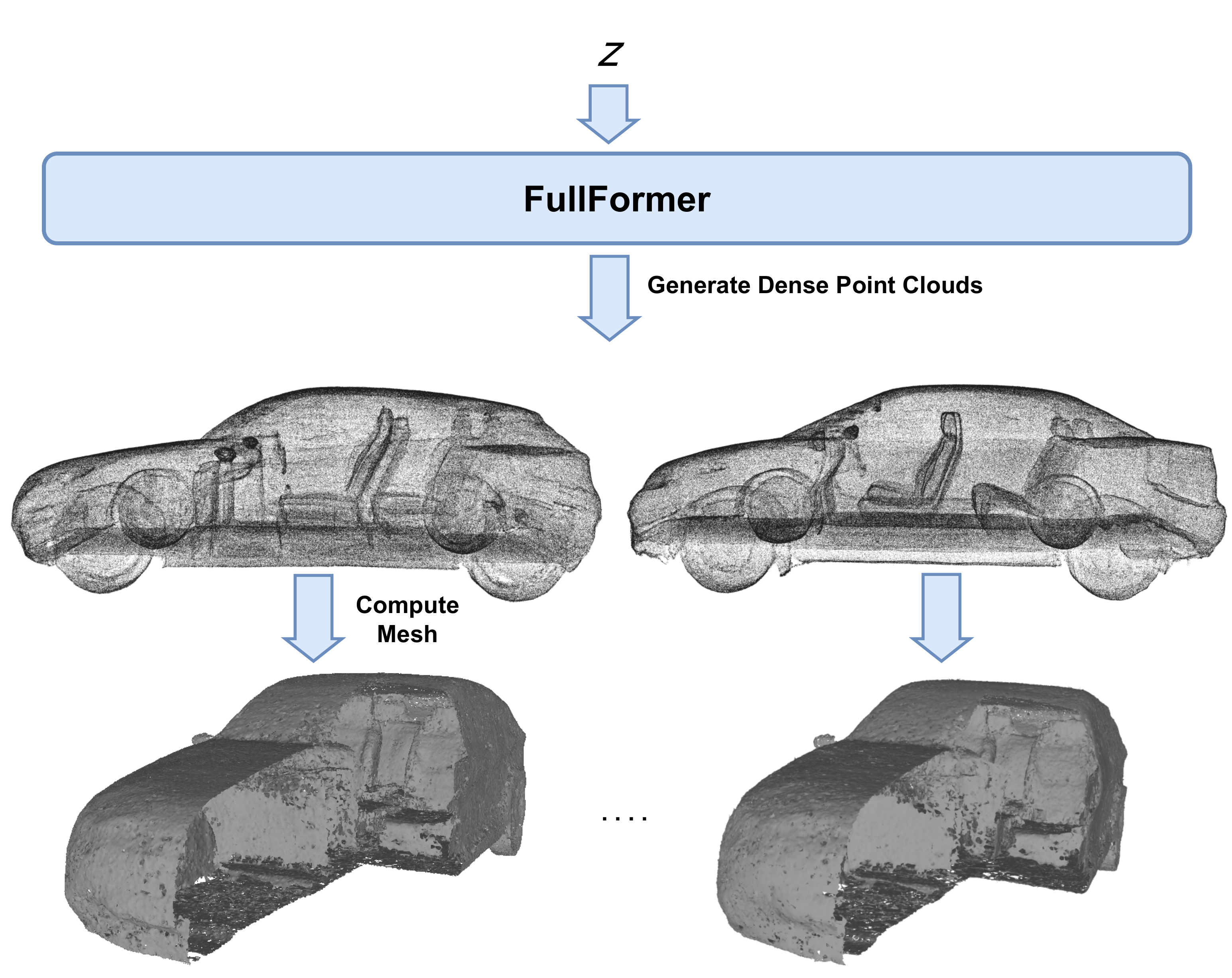}
   \caption{This paper addresses generating 3D objects with rich internal geometric details.}
  \label{fig:teaser}
  \vspace{-1.5em}
  
\end{figure}
Implicit representations for 3D shapes can be categorized into two types. The first type represents the outer surface of a 3D shape as occupancy grids or distance fields. Occupancy networks \cite{occupancy} define the surface as a continuous decision boundary of a deep neural network classifier whereas DeepSDF \cite{DeepSDF} represents a 3D surface using a signed distance field (SDF). A significant benefit of an SDF is easy extraction of the surface registered as a point cloud using the marching cubes algorithm \cite{lorensen1987marching}. However, many implicit neural networks require 3D shapes to be watertight which are often not readily available. Atzmon \etal~\cite{atzmon2020sal} propose a sign agnostic loss function to learn an SDF from non-watertight data; however, their model requires careful initialization of the neural network parameters and often misses thin structures. Another drawback of SDFs stems from their inherent nature \ie 3D shapes are modeled as inside and outside. Therefore multiple nested surfaces inside a 3D shape can not be represented due to only two states of distances in an SDF, either positive or negative and 0 or 1 in occupancy networks. 


Unsigned distance fields (UDFs) are another type of implicit representation whereby a 3D shape is delineated through a function that predicts the unsigned distance of a given point in space to the nearest surface of the 3D shape. This representation is capable of encoding multiple layers of internal 3D structures since distance values are not limited to only capturing inside or outside. However, extracting a surface from a UDF to a tractable datatype such as point clouds is non-trivial. The standard marching cube algorithm \cite{lorensen1987marching} cannot be used, as finding a zero-level set by detecting the flips between inside and outside is not possible with UDFs. Chibane et al.~\cite{chibane2020neural} provided algorithms to extract point clouds comprising internal geometries from UDFs. They have further demonstrated the use of UDFs on the task of shape reconstruction. However, shape completion/synthesis or novel shape generation with UDFs remains unexplored. In this paper, we present an approach, which leverages UDFs capability to represent nested 3D shapes to learn and generate rich internal structures, while ensuring the high quality and diversity of the 3D shapes.

Learning representations of complex shapes requires the encoding of distant shape contexts. This is especially true when shapes with internal structures are considered, \ie local shape context is not sufficient to model long-range relationships for example between the overall height of a car and the shape or tilting of its seats. To facilitate the encoding of relationships at varying spatial distances, transformer-based models that leverage the self-attention mechanism are the method of choice~\cite{Attention,dosovitskiy2020image}. Transformers are proven to be effective in modeling data distributions and generating realistic samples in image generation and 3D shape completion tasks \cite{esser2021taming,yan2022shapeformer}. Unfortunately, transformers can not directly learn from UDF representations since they rely on discrete token representations. Leveraging the advantages of transformers for shape generation with internal structure is therefore non-trivial. 

In this paper, we present a way to properly learn to generate 3D shapes with internal details while modeling long-range shape dependencies. This effectively integrates transformer-based shape learning with UDFs. We thereby exploit the fact that fine surface details can be locally encoded whereas global structures depend on large spatial contexts.
Thus, we first employ a convolutional neural network-based implicit function to encode the continuous latent representation of 3D shapes locally. We then make use of vector quantization to discretize the continuous locally encoded shape information into a sequence of discrete tokens. Finally, a latent transformer model is employed to learn the global long-range dependencies on the basis of these discrete tokens to generate novel discrete latent shape representations. This encoded 3D shape information is fed into the decoder to predict UDFs. 3D shapes are retrieved in the form of dense point clouds from the generated UDFs using the dense point cloud algorithm mentioned in \cite{chibane2020neural}. 

In summary, our contributions are as follows: 
\begin{itemize}
    \item We propose an implicit neural network-based generative framework for generating 3D shapes with nested geometries, \ie internal details.
    
    \item Our generative model can learn from both watertight and non-watertight 3D data.
    
    \item We carefully curate a new dataset of car shapes, with internal geometries, from ShapeNet and make it publicly available.
    
    \item We demonstrate that our method outperforms previous state-of-the-art and achieves superior qualitative and quantitative point cloud generation results.

\end{itemize}

\section{Related Work}

\paragraph{Generative Adversarial Networks} A standard generative model used in computer vision applications is the generative adversarial network (GAN)\cite{goodfellow2014generative}. Recent works \cite{chen2019learning,kleineberg2020adversarial} have shown 3D shape generation combining implicit neural networks and generative adversarial networks. However, the quality of output suffers from mode collapse and catastrophic forgetting due to the instability of GAN training \cite{modelcollapse,thanh2020catastrophic}.

\paragraph{Score-based Models} Another form of generative models are denoising diffusion probabilistic models, also known as score matching models \cite{hyvarinen2005estimation, ho2020denoising,song2019generative,song2020score}. The main principle of these models is that they model the gradient of the log probability density function with respect to the real sample. This process is referred to as score matching. These models have achieved state-of-the-art in many downstream tasks such as super-resolution, and generation \cite{saharia2021image,cai2020learning,chen2020wavegrad,zeng2022lion}. However, they are slow at inference time, limiting their usage in real-time applications.

\paragraph{Likelihood-based Models}
Variational auto-encoders (VAEs) and autoregressive models (ARs) are two commonly known likelihood-based models and they both aim to learn a probability distribution over the input data. While VAEs are computationally efficient and fast at inference time, their generation quality is often inferior compared to that of GANs\cite{kingma2013auto,pmlr-v32-rezende14}. Conversely, autoregressive models (ARs) can represent data distribution with high fidelity but generate samples slowly \cite{oord2016conditional,razavi2019,parmar2018image,chen2020generative}. To overcome the limitations of these two models, hybrid models combining the autoregressive transformer models and vector quantized VAEs have been proposed to generate high-resolution realistic images \cite{esser2021taming}. Our proposed method builds upon this hybrid model setup and focuses on generating 3D shapes with internal structures.
Our approach is related to ShapeFormer \cite{yan2022shapeformer}, which employs a latent transformer architecture to learn from compact and discretely encoded sequences that approximate 3D shapes, specifically for 3D shape completion utilizing signed distance functions (SDFs). However, they do not tackle the task of unconditional shape generation. Moreover, they employ a local pooled PointNet model \cite{PointNet} for feature extraction, which can limit the expressiveness of the feature embeddings. In contrast, we demonstrate that incorporating locality inductive biases, as in CNNs, in extracted features allows for tractable feature embeddings. Therefore, we opt for using an IF-Net-based \cite{IF-Nets} encoder. While their approach is restricted to performing shape completion exclusively on watertight 3D models, our method offers the ability to generate novel shapes with internal structures and is not constrained by watertight-only models.

\paragraph{Implicit Neural Generative Models}

In recent years, neural implicit networks have gained significant attention for their efficacy in 3D representational learning \cite{park2019deepsdf,mescheder2019occupancy,atzmon2020sal,peng2020convolutional,sitzmann2020implicit,sarmad2022photo,zheng2022sdf,hui2022neural,SPAGHETTI}. While several models have explored implicit representation for 3D surface reconstruction, only a few have used it for 3D model generation \cite{zeng2022lion,SPAGHETTI}. In general, this type of neural representation encapsulates a 3D surface by taking a spatial coordinate value as input and outputs a parameter, ones or zeros for points inside or outside the surface~\cite{mescheder2019occupancy} or a signed distance from the surface~\cite{park2019deepsdf}. However, as mentioned before, these representations do not preserve the internal geometry of 3D shapes. Recently, NDF~\cite{chibane2020neural} has demonstrated that UDFs are capable of representing inner details within 3D models. In this context, we propose a deep implicit generative framework that utilizes UDFs to generate high-quality 3D models with internal geometric structures. Our work highlights the potential of UDFs in generating rich 3D models. This has significant implications for various applications, such as product design, robotics, CAD designs, and medical imaging, whereby internal geometries are crucial for accurate modeling and simulation.


\section{Method}
\begin{figure*}[ht]
\centering
\includegraphics[width=180mm]{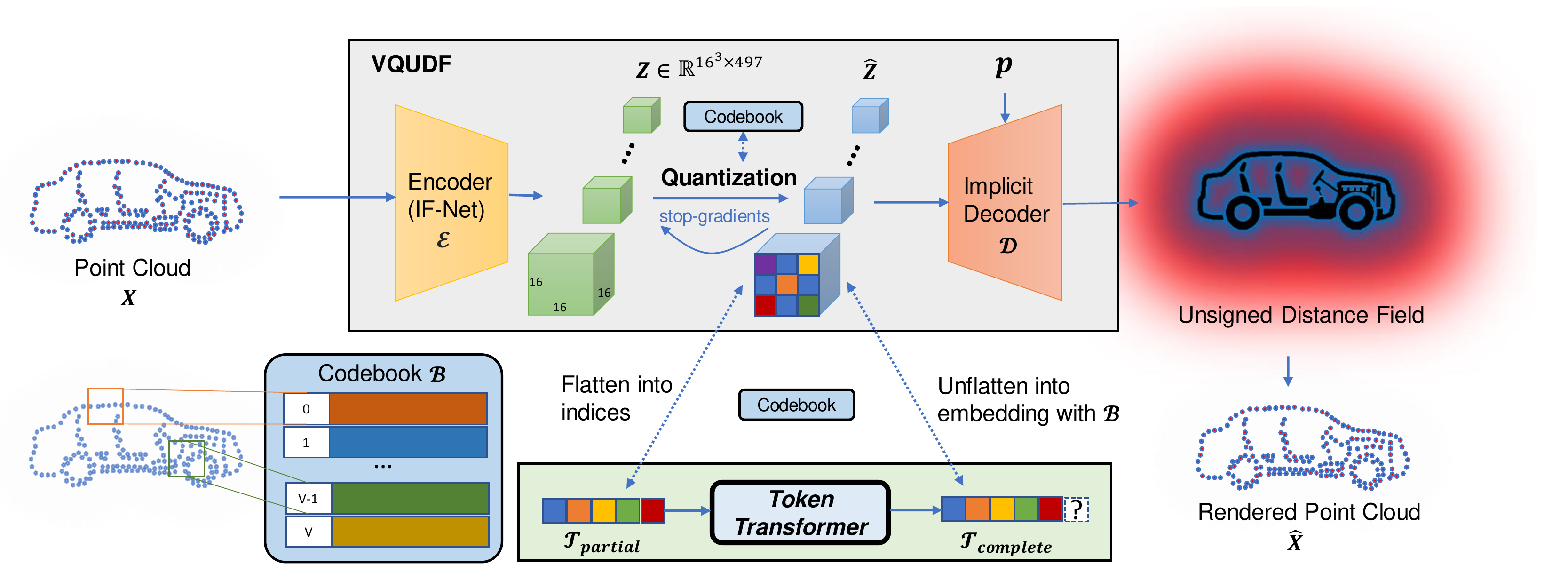}
\caption{
\textbf{Approach:} Key ingredients of our pipeline are vector quantized autoencoder, unsigned distance field (UDF), and latent transformer. The first stage is learning VQUDF which is a vector quantized autoencoder model that takes voxelized point clouds as input to a CNN-based encoder and utilizes an implicit decoder to output a UDF of the 3D shape. UDF ensures rich internal details are retained in a continuous data representation. Latent codes from the learned VQUDF are used to train an autoregressive transformer. This transformer learns to generate novel latent codes at test time. An implicit decoder then decodes generated latent codes to output a UDF. A 3D shape is then rendered from the UDF as a more tractable data format such as a point cloud.}
\label{fig:2-step}
\vspace{-1.0em}
\end{figure*}
 The objective of this work is to leverage the representational power of unsigned distance fields (UDF) in order to implicitly model 3D shapes whilst retaining their internal geometric details. To achieve this goal, we utilize the learning capabilities of transformers and incorporate UDF-based implicit function learning to develop an autoregressive generative model capable of generating 3D shapes with internal structures. Previous research works \cite{esser2021taming,yan2022shapeformer} have demonstrated the expressive power of transformers in capturing long-range dependencies in the input data. However, complexity increases considerably with the sequence length \cite{Attention}. This problem is exasperated when the data representation is a dense 3D model. Therefore, instead of representing a 3D model as voxels, point clouds, or discrete patches directly, we learn a compact and discrete representation whereby a shape is encoded using a codebook of context-rich parts. This allows a transformer to capture long-range interactions between these contextual parts and effectively model the distributions over the full shapes. At inference time, generated latent codes are decoded into a UDF using an implicit decoder. UDFs are then rendered into point clouds using a dense point cloud algorithm provided by Chibane et al. \cite{chibane2020neural}. Figure \ref{fig:2-step} details the complete framework of our approach. 

Our method can be sectioned into two parts. First, we describe a form of an autoencoder, namely Vector Quantized Unsigned Distance Field (VQUDF), which learns a context-rich codebook, as detailed in Sec.~\ref{VQUDF Desc.}. Then we present the latent transformer architecture as a generative model capable of producing novel shapes, as outlined in Sec.~\ref{transformer desc.}.

\subsection{Sequential Encoding with VQUDF}\label{VQUDF Desc.}

A 3D shape is represented as a point cloud input denoted by $\textbf{X} \in \mathbb{R}^{N \times 3}$. To harness the power of transformers in the generation, we encode $\textbf{X}$ into a discrete \textit{sequence} of tokens. This discrete \textit{sequence} must encapsulate the complete geometric information of the 3D shape. Inspired by ideas from \cite{van2017neural,esser2021taming}, we formalize the encoder, codebook, and decoder architecture for generating 3D shapes with internal geometry using UDFs.

\paragraph*{Encoder}
To generate 3D shapes with internal structures using transformers, we require a compact and discrete representation of the input shape that maintains high geometric resolution. The input to our encoder is sparse voxelized point cloud defining a 3D shape. When dealing with voxel data representations, capturing local spatial context is essential since the correlation between neighboring voxels significantly impacts the overall shape of the object. CNNs  are well-suited for capturing prior inductive bias of strong spatial locality within the images \cite{d2021convit}. By incorporating local priors from CNNs, we can effectively capture the spatial context of the input data and encode it into a compact feature grid utilizing ideas from neural discrete representation learning \cite{van2017neural}. To achieve this, the first step is to employ a CNN-based feature extractor $\mathcal{E}$ called IF-Net \cite{chibane2020neural}. IF-Net takes a sparse voxelized point cloud $\textbf{X}$ and maps it to a set of \textit{multi-scale} grid of deep features $\textbf{F}_1,...,\textbf{F}_m$ s.t.~$\textbf{F}_k \in \mathcal{F}^{K^3}_k$ and $\mathcal{F}_k \in \mathbb{R}^{c}$. Note that the resolution $K$ reduces, and the number of channels $c$ increases as $k$ increases. For tractability, we interpolate feature grids $\textbf{F}_1,...,\textbf{F}_{m-1}$ to the scale of final feature grid $\textbf{F}_m$ using trilinear interpolation. This provides us with a good trade-off between model complexity and shape details. A concatenation of $\textbf{F}_1,...,\textbf{F}_m$ along the channel dimension results in a compact feature grid $\textbf{Z} \in \mathbb{R}^{K^3 \times C}$. Note that $\textbf{Z}$ is a continuous latent feature representation. 

 
\paragraph*{Quantization}

A discrete description of the world can aid learning by compressing information in many domains, such as language or images \cite{van2017neural, mnih2014neural, chen2016variational}. We posit that 3D models are no exception and can greatly benefit from discrete representations. In addition, to utilize the generative transformer model, the input shape is preferably a discrete \textit{sequence}. Therefore, we employ vector quantization to transform the continuous latent feature representation $\textbf{Z}$ into a sequence of tokens $\mathcal{T}$ using a learned codebook $\mathcal{B}$ of context-rich codes $\mathcal{B} = \{\textbf{b}_i\}_{i=1}^{V} \subset \mathbb{R}^{n_z}$ where $n_z$ is  the length $K \times C$ of a code. Following a row-major ordering \cite{esser2021taming}, each feature slice $\textbf{z}_i \in \textbf{Z}$ is clamped to the nearest code in the codebook $\mathcal{B}$ using equation \ref{clamping}, fig. \ref{fig:2-step}, which results in a quantized feature grid $\hat{\textbf{Z}}$.
\begin{equation}
t_i = \text{argmin}_{ j \in \{1,..,V\}} \|\textbf{z}_i - \textbf{b}_j\| 
\label{clamping}
\end{equation}
A sequence of tokens $\mathcal{T}$ is then defined as the ordered set of indices $(t_i) \forall{i \in \{1,..,|\mathcal{T}|\}}$.

\paragraph*{Decoder}
As stated earlier, we aspire to learn an implicit representation of shapes to benefit from properties of such models, for example, no watertight shape restrictions, arbitrary resolution, and encoding internal structures. To achieve this, we train a decoder to output an unsigned distance field $\text{UDF}(\textbf{p}, \mathcal{S}) = \text{min}_{\textbf{q}\in\mathcal{S}}\|\textbf{p}-\textbf{q}\|$ which is a function that approximates the unsigned distances between the sample points $\textbf{p}$ and the surface of the shape $\mathcal{S}$. Formally, the decoder is defined as a neural function $\mathcal{D}(\hat{\textbf{Z}}, \textbf{p}): \mathbb{R}^{K^3 \times C} \times \mathbb{R}^3 \mapsto \mathbb{R}^+$ that regresses the UDF from a set of point $\textbf{p}$ conditioned on the latent discrete feature grid $\hat{\textbf{Z}}$. The dense point cloud algorithm provided by Chibane et al. \cite{chibane2020implicit} is used further to convert UDF to a final point cloud denoted by $\hat{\textbf{X}}$.

\paragraph*{Training VQUDF} \label{I stage training}

The training process involves learning the encoder $\mathcal{E}$, codebook $\mathcal{B}$, and the decoder $\mathcal{D}$ simultaneously. The overall loss function is denoted in equation \eqref{vqudf_loss}.
%
\begin{align}
&\mathcal{L}_{\text{VQUDF}}(\mathcal{E},\mathcal{B},\mathcal{D}) = \nonumber\\
&\parallel \text{UDF}(\textbf{p},\mathcal{S}) - \text{UDF}_{gt}(\textbf{p},\mathcal{S}) \parallel_2^2 + \mathcal{L}_c 
\label{vqudf_loss}
 \end{align}
The first term denotes the reconstruction loss, which is computed as the difference between predicted and ground truth UDFs. This method is different from the commonly utilized approach of computing loss between predicted and true point clouds. The second term $\mathcal{L}_c$ denotes the commitment loss in equation \eqref{comm_loss}.
\begin{equation}
\mathcal{L}_c =\parallel \text{sg}[\mathcal{E}(\textbf{X})] - \hat{\textbf{Z}} \parallel^2_2 + \parallel \text{sg}[\hat{\textbf{Z}}] - \mathcal{E}(\textbf{X})\parallel^2_2
\label{comm_loss}
 \end{equation}
%
Different from vanilla NDF training, our pipeline has a non-differentiable quantization operation. Following previous works \cite{Gradient_approach, van2017neural}, we utilize a straight-through gradient estimator to circumvent this problem. Under this approach, gradients are simply copied over from the decoder to the encoder. This method ensures joint training of the codebook, the encoder, and the decoder.

\subsection{Generating a Sequence of Latent Vectors}\label{transformer desc.}

\paragraph*{Latent Transformer}    

Transformers have shown tremendous performance in generating images by modeling them as a sequence of tokens and learning to generate such sequences \cite{Transimage,Sequenceimage}. Transformers are unconstrained by the locality bias of CNNs allowing them to capture long-range dependencies in images. 3D models with internal structures also exhibit long-range dependencies, for example, the number and shape of seats in a car depend on the body being either a sedan or a sports car. Previous works \cite{zhao2021point,guo2021pct,xiang2021snowflakenet} have successfully demonstrated capturing these dependencies using transformers for 3D models. We represent 3D shapes as a sequence of tokens $\mathcal{T} = (t_1,...,t_{|\mathcal{T}|})$ resulting from our trained VQUDF framework. Recall that each token $t_i$ is an index of the closest codebook latent embedding  to the continuous latent feature grid. The generation of shapes is modeled as an autoregressive prediction of these indices. A transformer learns to predict the distribution of the next indices given prior ones. The likelihood of the complete sequence $\mathcal{T}$ is described as $p(\mathcal{T}) = \prod_{i=1}^{|\mathcal{T}|} p(t_i|t_{1...i-1})$.

\paragraph*{Transformer Training }
The generation of latent codes as a sequence of tokens using transformers is highlighted in Fig.~\ref{fig:2-step}. The learned weights of the trained VQUDF autoencoder are frozen before the training of the transformer. VQUDF is first used to create a training dataset of 3D shape latent embeddings. These latent embeddings are used in the training of the transformer. The training objective for generation is maximizing the log-likelihood of tokens in a randomly sampled sequence to represent the 3D shape $p(\mathcal{T})$:
\begin{align}
\mathcal{L}_{\text{Transformer}}  = \mathbb{E}_{x \sim p(x)}[-\text{log} \: p(\mathcal{T})] 
\label{Log-like}
\end{align}
\par\noindent
After training, this model starts with the [START] token and predicts the next indices forming a complete sequence $\mathcal{T}$ until a [END] token is predicted.
By mapping indices in the sequence $\mathcal{T}$ back to the corresponding codebook entries, a discrete latent feature grid $\hat{\textbf{Z}}$ is recovered. The 3D shape is then reconstructed using the implicit decoder $\mathcal{D}$, which results in a UDF. We use the dense point cloud extraction algorithm proposed in \cite{chibane2020neural} to extract the generated 3D shape $\hat{\textbf{X}}$ as a point cloud.


\section{Experiments}
This section thoroughly evaluates our proposed approach and demonstrates its effectiveness in generating high-quality shapes with internal structure and details. We compare our point cloud generation results against multiple SOTA point cloud generation baselines and present good qualitative and quantitative results for the shape generation task.

\subsection{Implementation Details}
We train our models in two stages. First, we train the VQUDF module, followed by a latent transformer module. For training, we utilize stock hardware comprising one Nvidia RTX Quadro GPU with 48GB of VRAM. All code is written in PyTorch \cite{NEURIPS2019_9015} whereby a portion is acquired from open repositories of \cite{chibane2020neural, esser2021taming}. For training both modules, we use a batch size of 1 and the Adam optimizer. For VQUDF training, we employ a learning rate of 1e-6 and ReLU activation, whereas the transformer's training uses a learning rate of 4.5e-6. Furthermore, the transformer has 12 layers and 8 attention heads. The length of the input sequence to the transformer model is set as 7952; the codebook size is 8192, with each codebook having a dimensionality of 512. Additional training details, including architectures of networks, are presented in the supplementary material.

\subsection{Datasets}
\begin{figure}[ht]
\begin{center}
\begin{tabular}{ c c c }
 \begin{subfigure}{0.14\textwidth}
      \includegraphics[trim={0.2cm 0.2cm 0.2cm 0.2cm},clip, width=\textwidth]{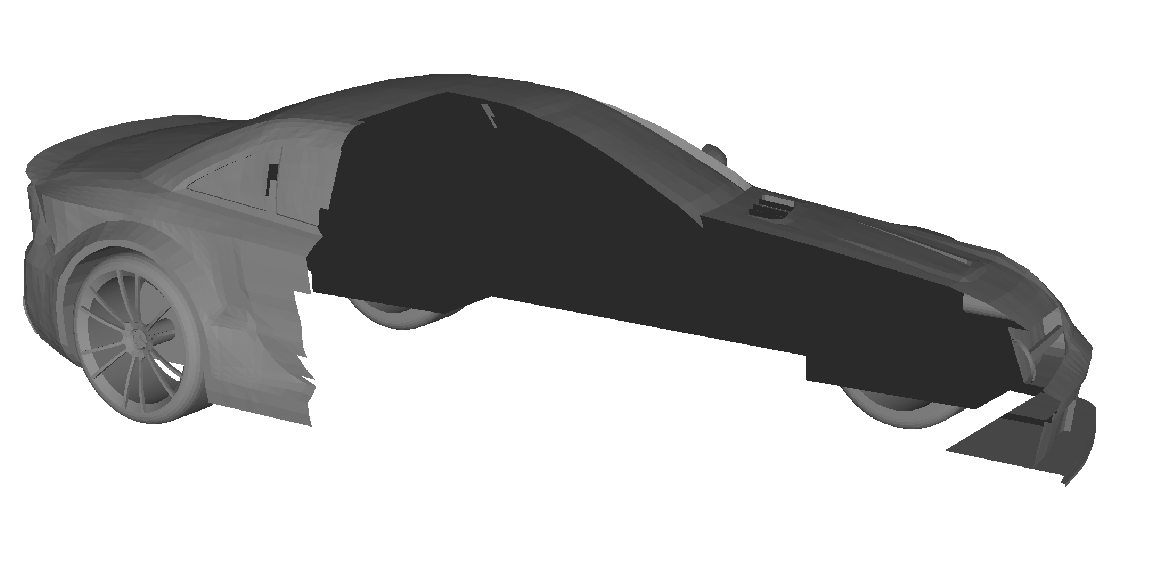}
    \end{subfigure}
     &  \begin{subfigure}{0.14\textwidth}
      \includegraphics[trim={0.2cm 0.2cm 0.2cm 0.2cm},clip, width=\textwidth]{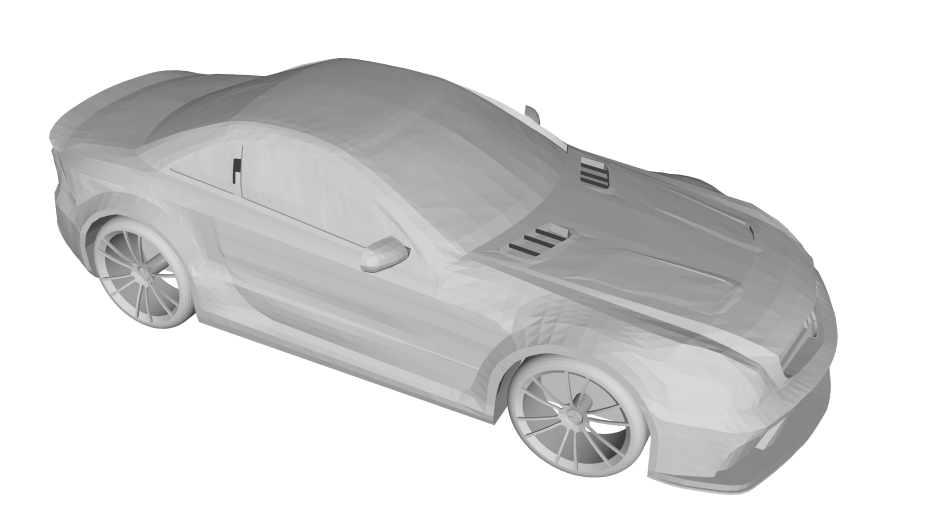}
    \end{subfigure} & 
    \begin{subfigure}{0.14\textwidth}
      \includegraphics[trim={0.2cm 0.2cm 0.2cm 0.2cm},clip,width=\textwidth]{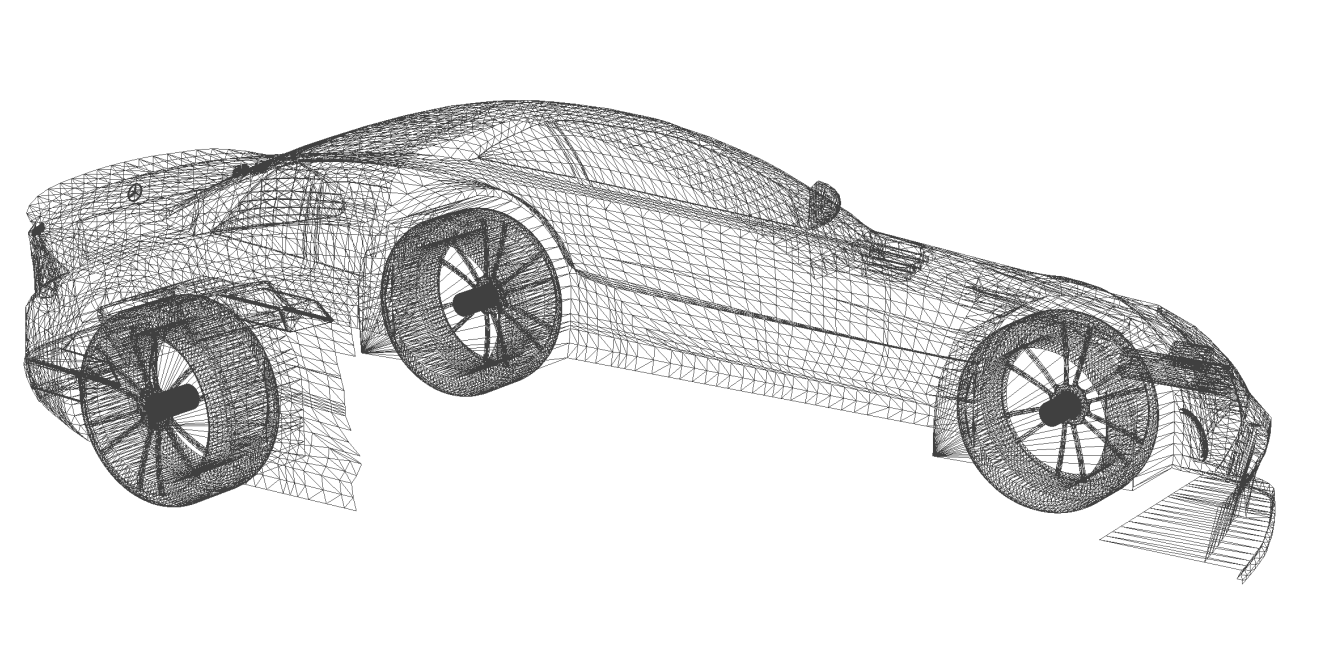}
    \end{subfigure}
\end{tabular}
\end{center}
\vspace{-1.5em}
\caption{\textit{ShapeNet:} Most samples have no internal details.}
\label{fig:badcars}
\vspace{-1.5em}
\end{figure}

Since our approach focuses on generating internal structures, we sought a dataset of 3D shapes having internal geometric details. Such datasets are scarce which led us to curate a new dataset of cars with realistic internal geometry. We call this dataset `Full Cars'. Other than this dataset, we utilize standard datasets from ShapeNetCore \cite{Shapenet:Dataset}. A detailed description of both datasets is provided in this section.
\paragraph{ShapeNet}
We use the ShapeNetCore v2 dataset with three categories: \textit{Airplanes}, \textit{Chairs}, and \textit{Cars}. The Cars object category of the ShapeNet dataset contains both open and closed shapes. However, some shapes are present with internal details. Most cars either include no internal structure or significantly degraded internal geometry, as shown in Fig. \ref{fig:badcars}.

\begin{figure}[ht]
\begin{center}
\begin{tabular}{ c c c }
 \begin{subfigure}{0.14\textwidth}
      \includegraphics[trim={0.2cm 0.2cm 0.2cm 0.2cm},clip, width=\textwidth]{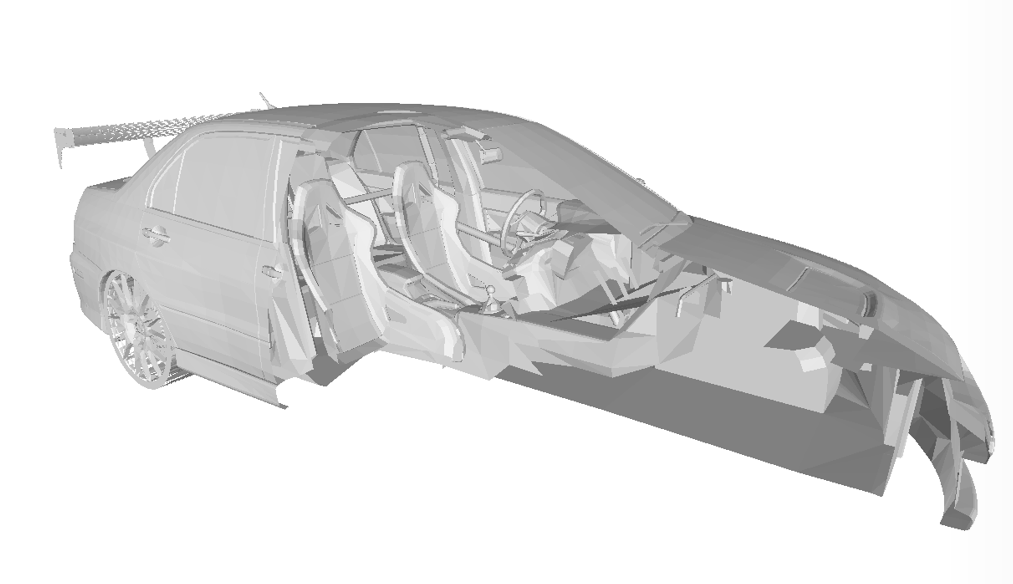}
    \end{subfigure}
     &  \begin{subfigure}{0.14\textwidth}
      \includegraphics[trim={0.2cm 0.2cm 0.2cm 0.2cm},clip, width=\textwidth]{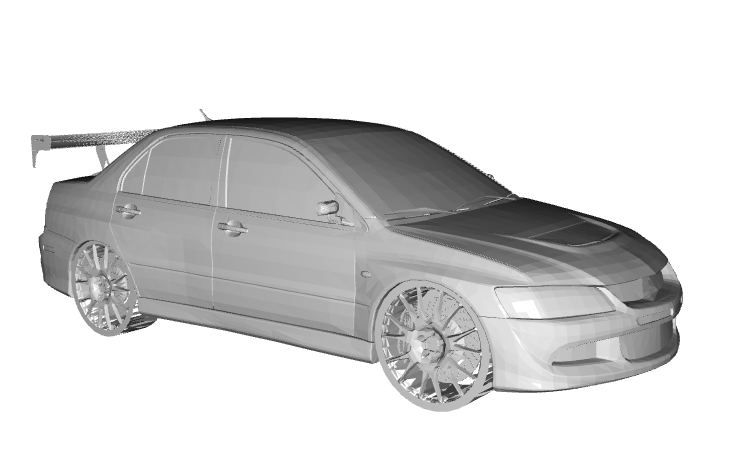}
    \end{subfigure} & 
    \begin{subfigure}{0.14\textwidth}
      \includegraphics[trim={0.2cm 0.2cm 0.2cm 0.2cm},clip,width=\textwidth]{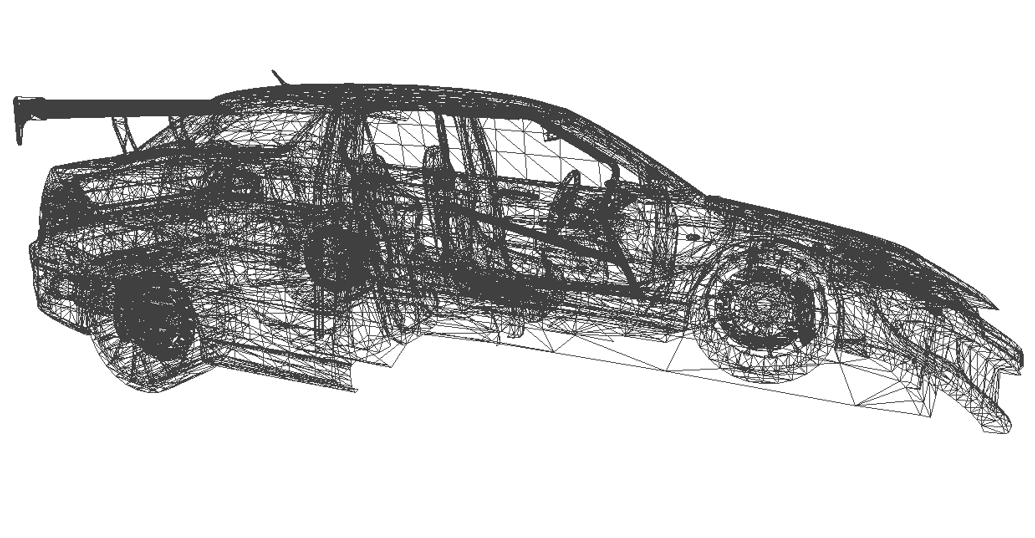}
    \end{subfigure}
\end{tabular}
\end{center}
\vspace{-1.5em}
\caption{\textit{Full Cars:} We curate a dataset of cars from ShapeNet that contains rich internal details.}\label{fig:fullcar}
\vspace{-1.5em}
\end{figure}

\paragraph{Full Cars}
This dataset is a subset of the ShapeNetcore v2 dataset of the `cars' category. The cars in this dataset exhibit rich internal details. We utilize Blender to filter out cars without internal structures. We select cars that have a significant number of points inside the outer shell to facilitate learning of insides. After curating, the final dataset contains 1602 models of full cars. These have been split into 1122, 320, and 160 shapes for training, validation, and test sets, respectively. Note that although this dataset comprises cars with internal structure, these models are non-watertight and therefore present a significant challenge for implicit representation and subsequent generation. A glimpse of the internal geometry in this dataset is shown in Fig.~\ref{fig:fullcar}.

\subsection{VQUDF Reconstruction Performance}
\begin{figure}[ht]
\begin{center}
\scalebox{2.0}{\begin{tabular}{@{}c@{}c@{}c@{}}

      &\includegraphics[trim={0.7cm 0.5cm 0.7cm 0.5cm},clip, width=0.14\textwidth]{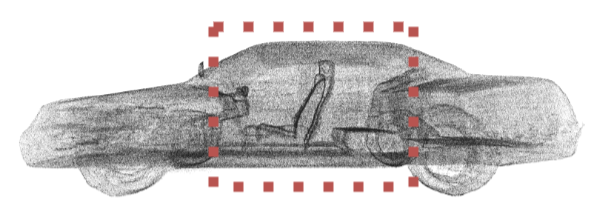}
     
     \\
     & \includegraphics[trim={0.5cm 0.5cm 0.7cm 0.5cm},clip,width=0.14\textwidth]{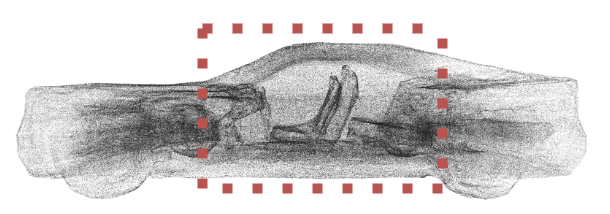}
    
    \\
    &  \includegraphics[trim={0.20cm 0.2cm 0.45cm 0.2cm},clip,width=0.14\textwidth]{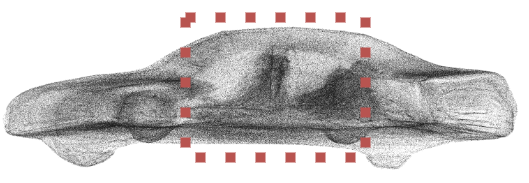}
        
\end{tabular}}
\end{center}
\vspace{-1.5em}
\caption{\textbf{Reconstruction Results:} Our model reconstruction results with different latent space resolutions $64^3$, $16^3$ and $8^3$ respectively (top to bottom).}
\vspace{-1.5em}
\label{fig:auroencoding}

\end{figure}

The input point cloud is sampled and voxelized before feeding into the VQUDF encoder. The number of points sampled from different datasets and voxel resolution during training of the VQUDF module are presented in Table~\ref{table:points and epochs}. Recall that the input 3D shape is encoded into a feature grid $\mathbf{\hat{Z}}$ where each channel comprises a feature block of dimension $K^3$. The quality of encoded information and generation capability depends on the dimensionality $K$ of the 3D latent feature grid $\hat{\textbf{Z}}$. Fig.\ref{fig:auroencoding} shows reconstruction results of the VQUDF module on the Full Cars dataset with different values of $K$ such that resolution of the 3D latent feature becomes $\hat{\textbf{Z}} \in \mathbb{R}^{64^3 \times C}$, $\hat{\textbf{Z}} \in \mathbb{R}^{16^3 \times C}$ and $\hat{\textbf{Z}} \in \mathbb{R}^{8^3 \times C}$ respectively, where $C$ is the number of channels. Note that the fidelity of internal geometries increases progressively with the dimensionality $K$ of $\hat{\textbf{Z}}$. However, increased $K$ results in a large quantized sequence length $\tau$ making transformer training difficult. Hence, a good trade-off between geometrical fidelity and memory footprint is achieved by selecting $\hat{\textbf{Z}} \in \mathbb{R}^{16^3 \times C}$ which is then processed into a tractable sequence of tokens to generate shapes with internal details.

\begin{table}[ht]
\captionsetup{justification=centering}
\begin{center}
 \caption[Number of points sampled and voxel resolution considered for training from different datasets.]{Number of points sampled and voxel resolution considered for VQUDF training for different datasets.}
 
    \scalebox{1}{
    \begin{tabular}{ccc} 
    \hline
     Dataset &  Points Sampled & Voxel resolution \\
     \hline\hline
     ShapeNet \textit{Cars} & 10000 & 256³ \\
    \hline
     ShapeNet \textit{Planes} & 5000 & 32³ \\ 
    \hline
     ShapeNet \textit{Chairs} & 4000 & 32³\\
    \hline
     Full Cars & 10000 & 256³\\
    \hline
    \end{tabular}}
   
    \label{table:points and epochs}
\end{center}
\vspace{-2.0em}
\end{table}

\begin{figure*}[ht]
\begin{center}
\newcommand{\rulesep}{\unskip\ \vrule\ }
\scalebox{0.99}{\begin{tabular}{@{}c@{}c@{}c@{}c@{} c@{} |c@{} c@{} c@{} c@{}}
      \includegraphics[trim={5cm 1cm 5cm 1cm},clip, width=0.1245\textwidth]{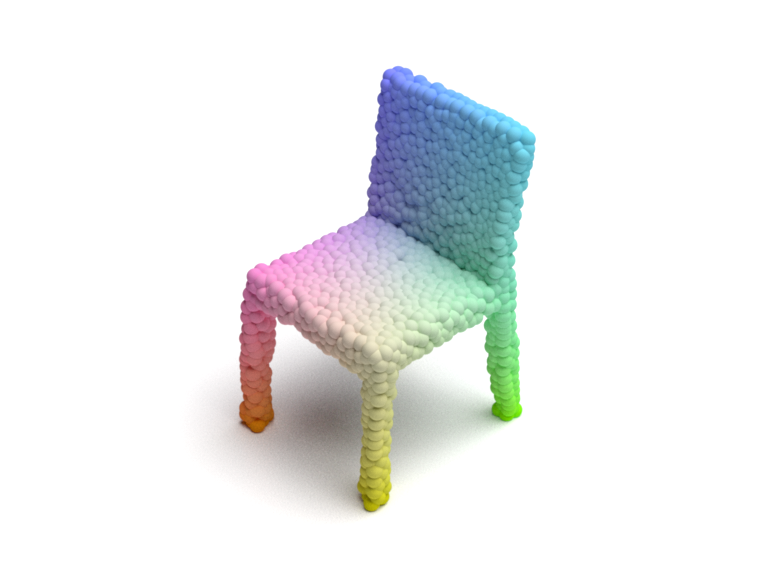}
     &  
      \includegraphics[trim={5cm 1cm 5cm 1cm},clip,width=0.1245\textwidth]{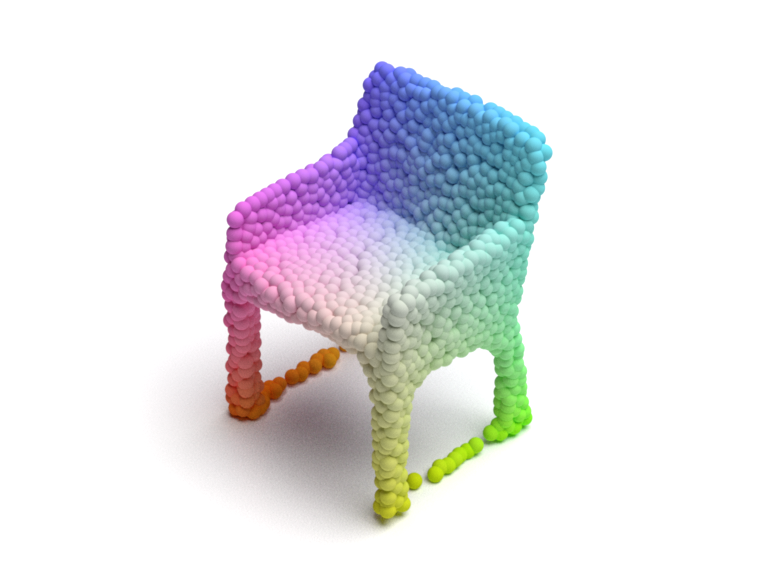}
   & 
    
      \includegraphics[trim={5cm 1cm 5cm 1cm},clip,width=0.1245\textwidth]{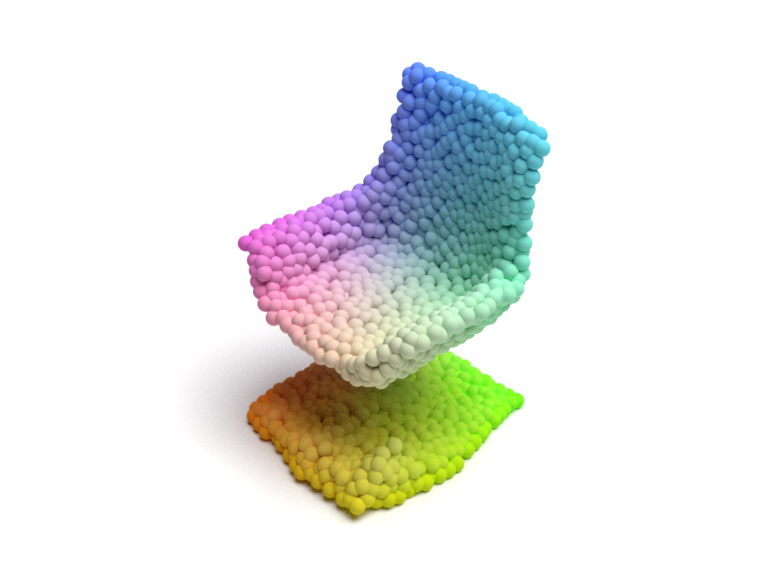}
       &

      \includegraphics[trim={4cm 0cm 5cm 1cm},clip,width=0.1245\textwidth]{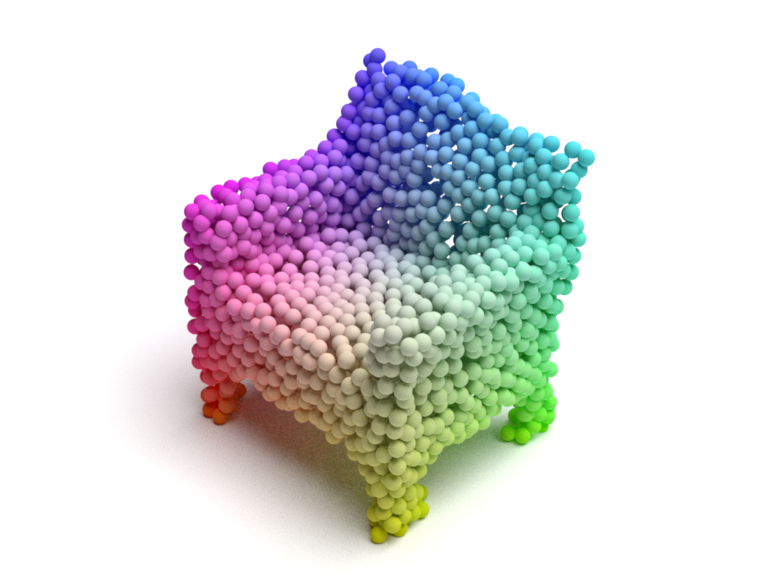}
      & 
      \includegraphics[trim={5cm 1cm 5cm 1cm},clip,width=0.1245\textwidth]{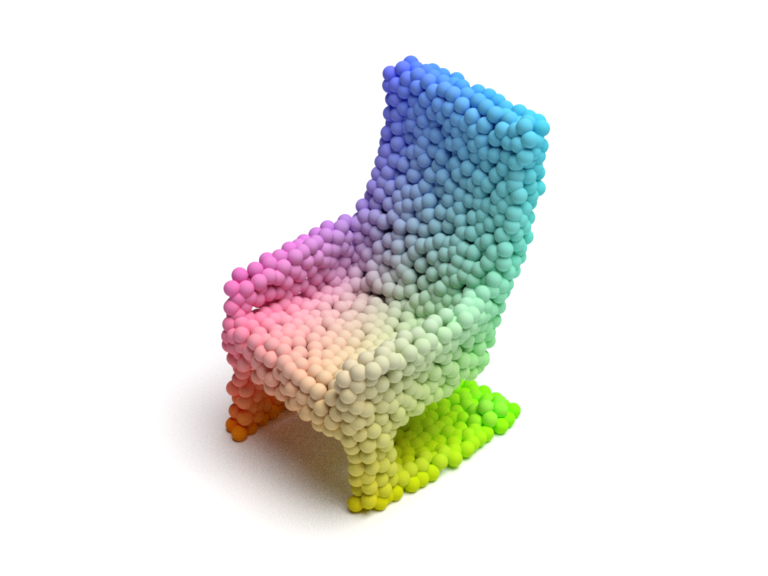}
      &
      \includegraphics[trim={5cm 1cm 5cm 1cm},clip,width=0.1245\textwidth]{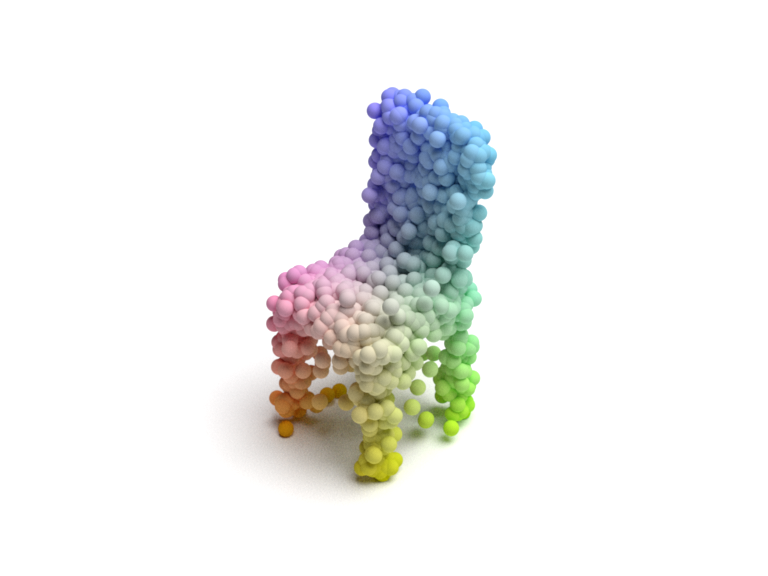}
      &
      \includegraphics[trim={5cm 1cm 5cm 1cm},clip,width=0.1245\textwidth]{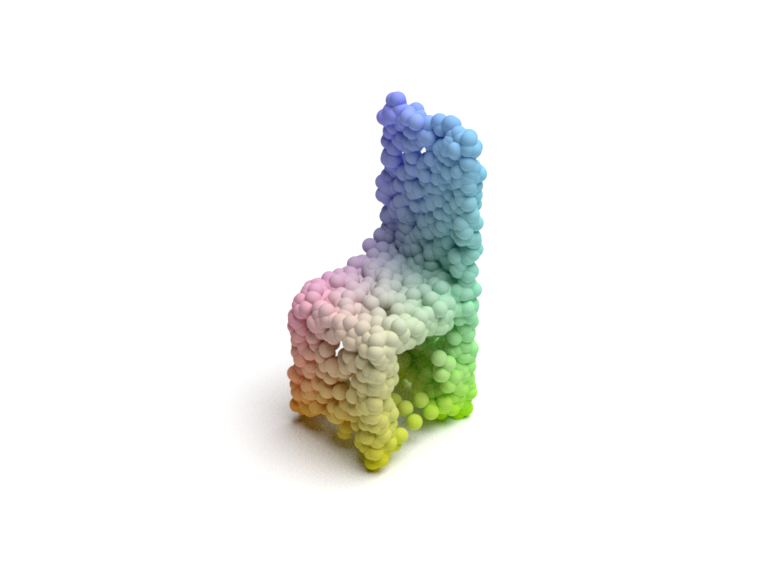}
      &
      \includegraphics[trim={5cm 1cm 5cm 1cm},clip,width=0.1245\textwidth]{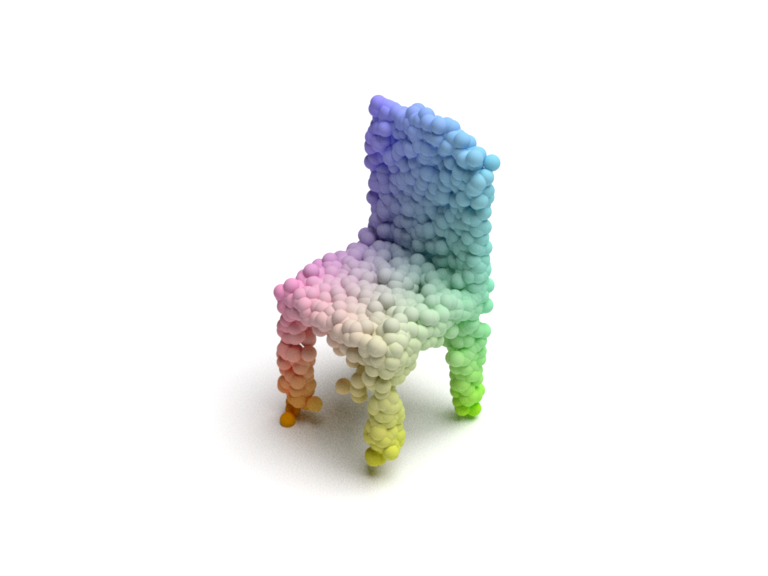}
    \\
      \includegraphics[trim={6cm 2cm 6cm 2cm},clip,width=0.1245\textwidth]{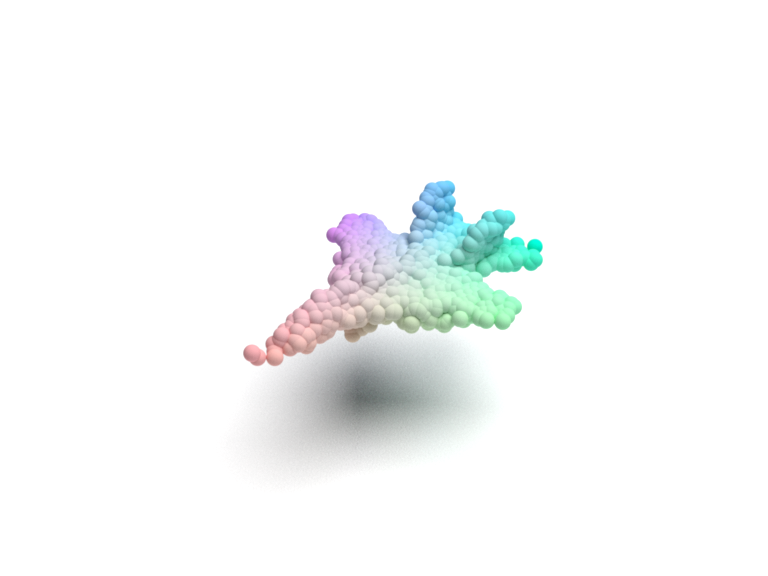}
 & 
      \includegraphics[trim={6cm 2cm 6cm 2cm},clip,width=0.1245\textwidth]{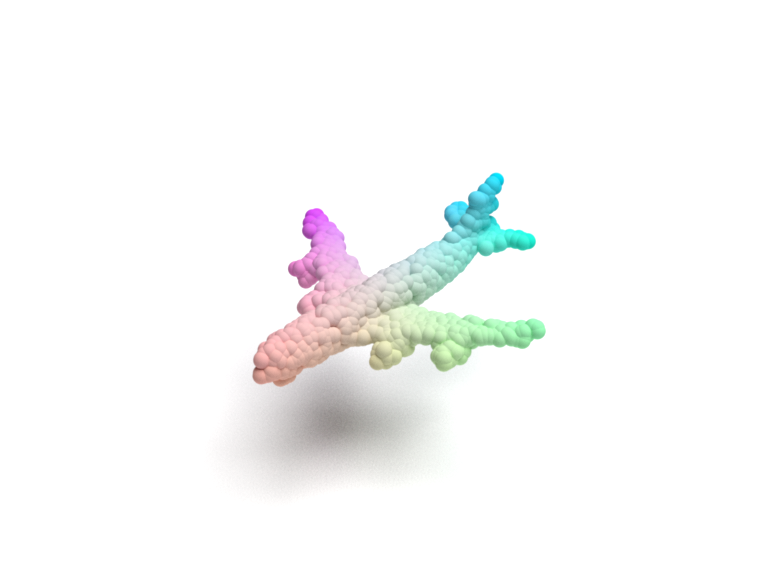}
     & 
      \includegraphics[trim={6cm 2cm 6cm 2cm},clip,width=0.1245\textwidth]{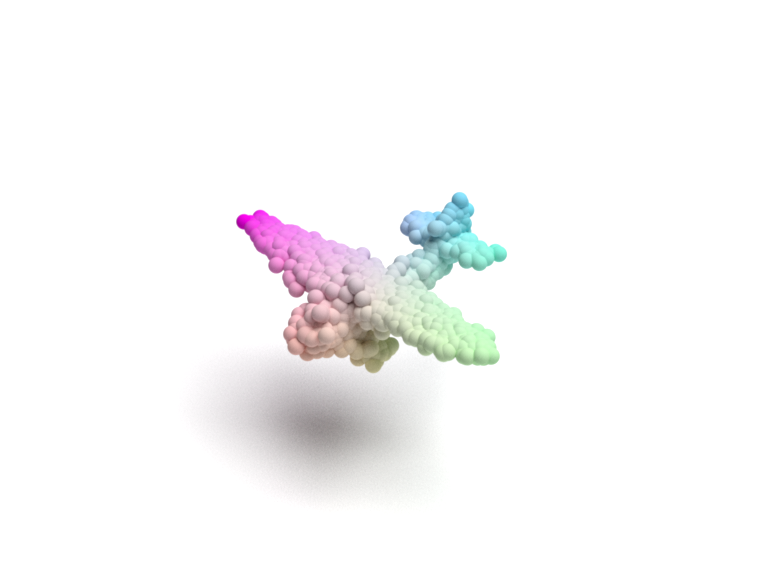}
      &
 
      \includegraphics[trim={6cm 2cm 6cm 2cm},clip,width=0.1245\textwidth]{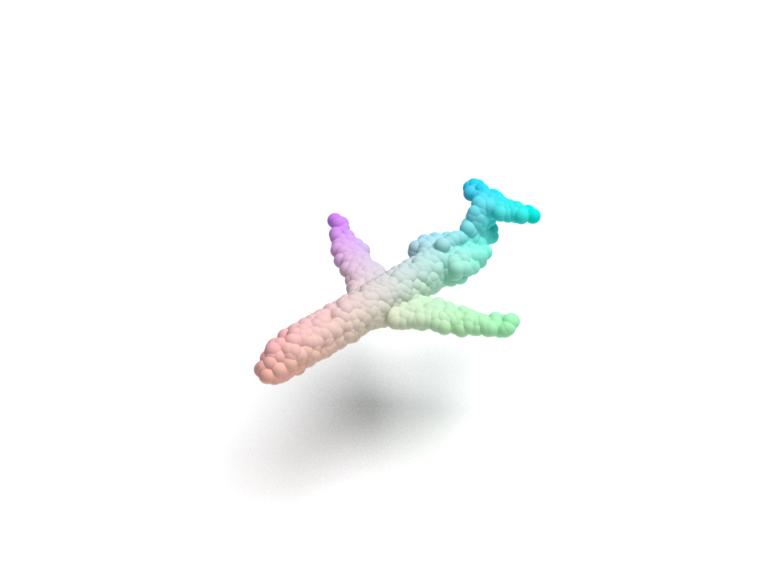}
       & 
      \includegraphics[trim={6cm 2cm 6cm 2cm},clip,width=0.1245\textwidth]{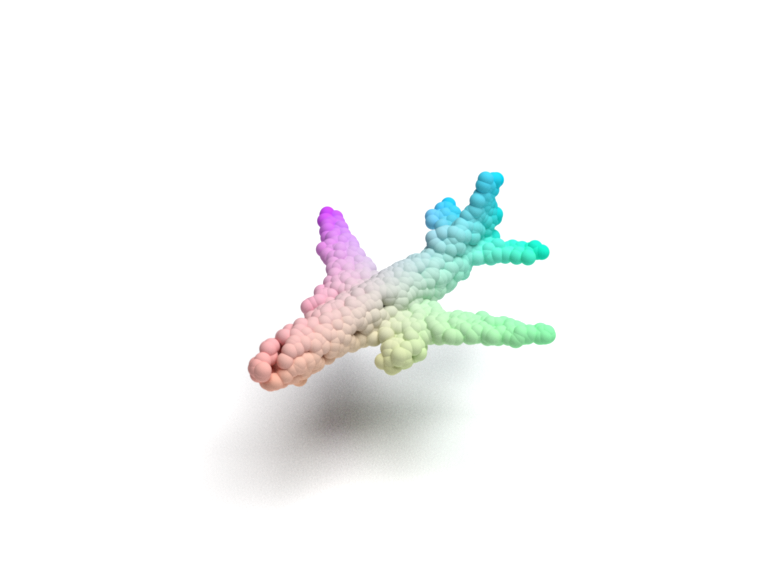}
       & 
      \includegraphics[trim={6cm 2cm 6cm 2cm},clip,width=0.1245\textwidth]{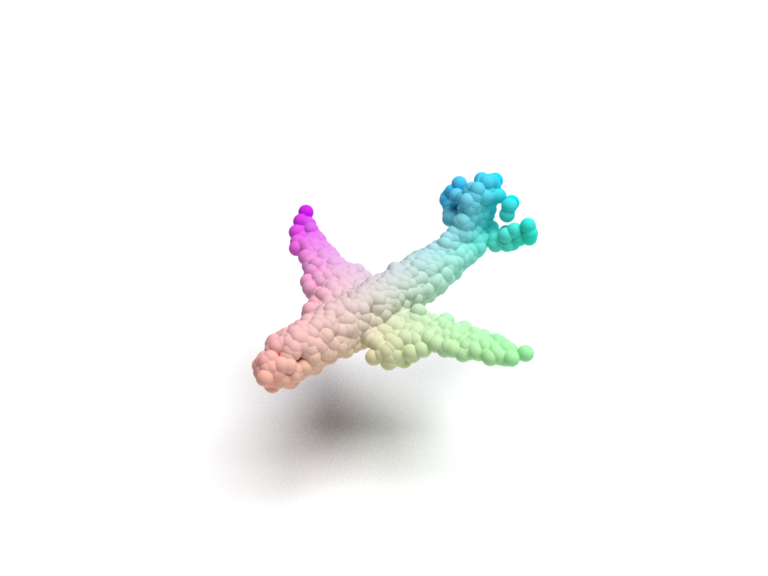}
      &
    \includegraphics[trim={6cm 2cm 6cm 2cm},clip,width=0.1245\textwidth]{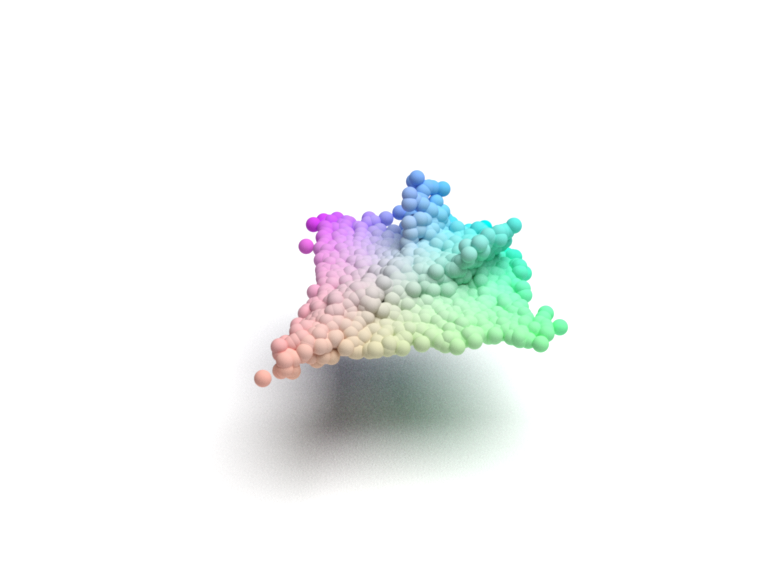}
    &
      \includegraphics[trim={6cm 2cm 6cm 2cm},clip,width=0.1245\textwidth]{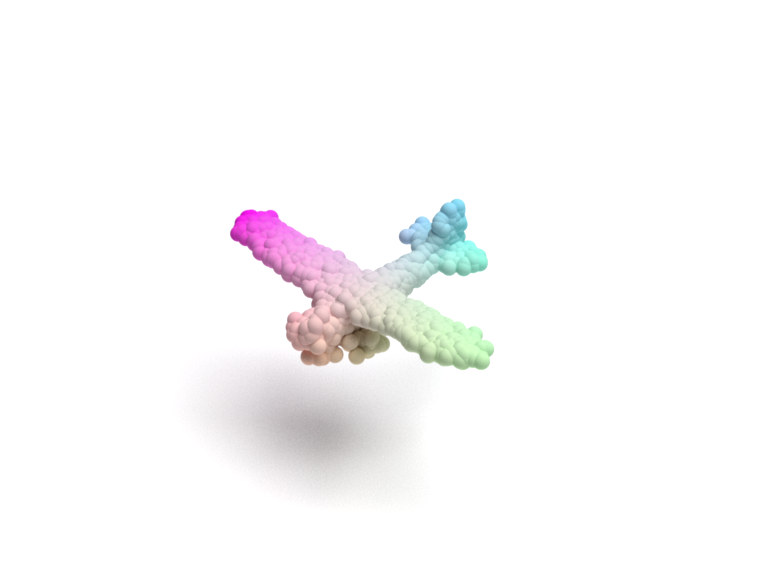}
      \\
      \multicolumn{5}{c}{FullFormer (Ours)}&GraphCNN-GAN&Diffusion&PointFlow
\end{tabular}}
\end{center}
\vspace{-1.5em}
\caption{\textbf{Outer Hull Generation:} Our models show high-quality point cloud generation results when trained on object categories of chairs, aeroplanes of ShapeNet dataset and visually improve over previous methods such as GraphCNN-GAN~\cite{valsesia2019learning}, Diffusion~\cite{luo2021diffusion} or PointFlow~\cite{yang2019pointflow}.}
\label{fig:outer}
\vspace{0em}

\end{figure*}

\subsection{Baseline} 
We use the following baselines which generate novel 3D point clouds to compare with our point cloud generation. The first baseline is Graph Convolution GAN \cite{valsesia2019learning}, which relies on standard GAN-based generation and employs localized operations in the form of graph convolutions to generate point clouds. Another baseline is Diffusion Model Luo et al. \cite{luo2021diffusion}, which employs denoising diffusion probabilistic models for the point cloud generation. Lastly, we also compare against Pointflow \cite{yang2019pointflow}, which utilizes normalizing flows for the point cloud generation. These models naturally carry the ability to learn inside details of 3D models, provided that they have been trained on datasets with internal structures. However, they do not utilize an implicit continuous representation to capture internal details. Therefore, these approaches are not only limited to a fixed number of points generation and resolutions but also their ability to model insides in predicted 3D shapes.


\subsection{Metrics}\label{metrics}
For quantitative evaluation, we use three different metrics following previous works. 

\paragraph{MMD} Minimum matching distance (MMD) indicates the faithfulness of generated samples with real data. A lower MMD indicates that generated samples are realistic towards ground truth samples.

\paragraph{COV} Diversity is an important aspect of generative models. A high coverage score (COV) indicates that the model does not suffer from mode collapse and has high sample diversity. 

\paragraph{JSD} Jenson-Shannon divergence (JSD) computes the symmetric similarity between distributions of generated samples and reference samples. A lower value of JSD is desirable. However, this metric is dependent on the selection of the reference set.


\begin{table*}[ht]
\centering
\caption{We quantitatively compare the results of our method with GraphCNN-GAN~\cite{valsesia2019learning}, Diffusion~\cite{luo2021diffusion} and PointFlow~\cite{yang2019pointflow}. We report minimum matching distance (MMD), coverage score (COV), and Jenson and Shannon divergence (JSD) for comparison. We use Chamfer distance (CD) for MMD and COV calculations. MMD scores are multiplied by
$10^{3}$ and JSD are multiplied by $10^{-1}$. Our proposed FullFormer improves consistently over all previous methods in terms of MMD and COV.}
\begin{tabular}{@{}lc@{\hspace{0.2cm}}c@{\hspace{0.2cm}}cc@{\hspace{0.2cm}}c@{\hspace{0.2cm}}cc@{\hspace{0.2cm}}c@{\hspace{0.2cm}}cc@{\hspace{0.2cm}}c@{\hspace{0.2cm}}c@{}}
\toprule
\multirow{2}{*}{\diagbox{Dataset}{Model}} &
\multicolumn{3}{c}{GraphCNN-GAN~\cite{valsesia2019learning}}&
\multicolumn{3}{c}{ Diffusion~\cite{luo2021diffusion}} &  \multicolumn{3}{c}{PointFLow~\cite{yang2019pointflow}} & \multicolumn{3}{c}{\textbf{Ours (FullFormer)}}\\
\cmidrule(r){2-4}\cmidrule(r){5-7}\cmidrule(r){8-10}\cmidrule(r){11-13}
    \multicolumn{1}{c}{} & MMD$\downarrow$ & COV$\uparrow$ & JSD$\downarrow$ & MMD$\downarrow$ & COV$\uparrow$  & JSD$\downarrow$ & MMD$\downarrow$ & COV$\uparrow$  & JSD$\downarrow$ & MMD$\downarrow$ & COV$\uparrow$ & JSD$\downarrow$ \\
\midrule
ShapeNet \textit{Cars}& 3.18 & 16 & 4.67  & 1.4 & 17.7 & \textbf{2.21} &  1.28 & 29.67 & 3.16 & \textbf{1.13} & \textbf{29.72} & 2.29 \\
%
ShapeNet \textit{Planes} &1.1  &31.09   &1.75   & 0.98 & 36.73 & \textbf{0.65}  & 1.41 & 35.87 & 1.06  & \textbf{0.92} & \textbf{37.37} & 0.83  \\

ShapeNet \textit{Chairs} &4.213   & 33.5   &1.24      &  \textbf{3.79} & 36.2 & \textbf{0.42}  & 4.19 & 33.23 & 0.82 & \textbf{3.79} & \textbf{37} & 1.06  \\

Full Cars  &2.32 & 20 & 3.81     & 1.24 & 21.23 & 2.83  & 1.18 & 24.85 & 3.39 & \textbf{0.93} & \textbf{25.07} & \textbf{2.72}  \\
\bottomrule
\end{tabular}
\label{tab:quant}
\end{table*}

\begin{figure*}[ht]
\begin{center}
\scalebox{1.965}{\begin{tabular}{@{}c@{}c@{}c@{}c@{}}

      \includegraphics[trim={0.5cm 0.5cm 0.5cm 0.5cm},clip, width=0.1245\textwidth]{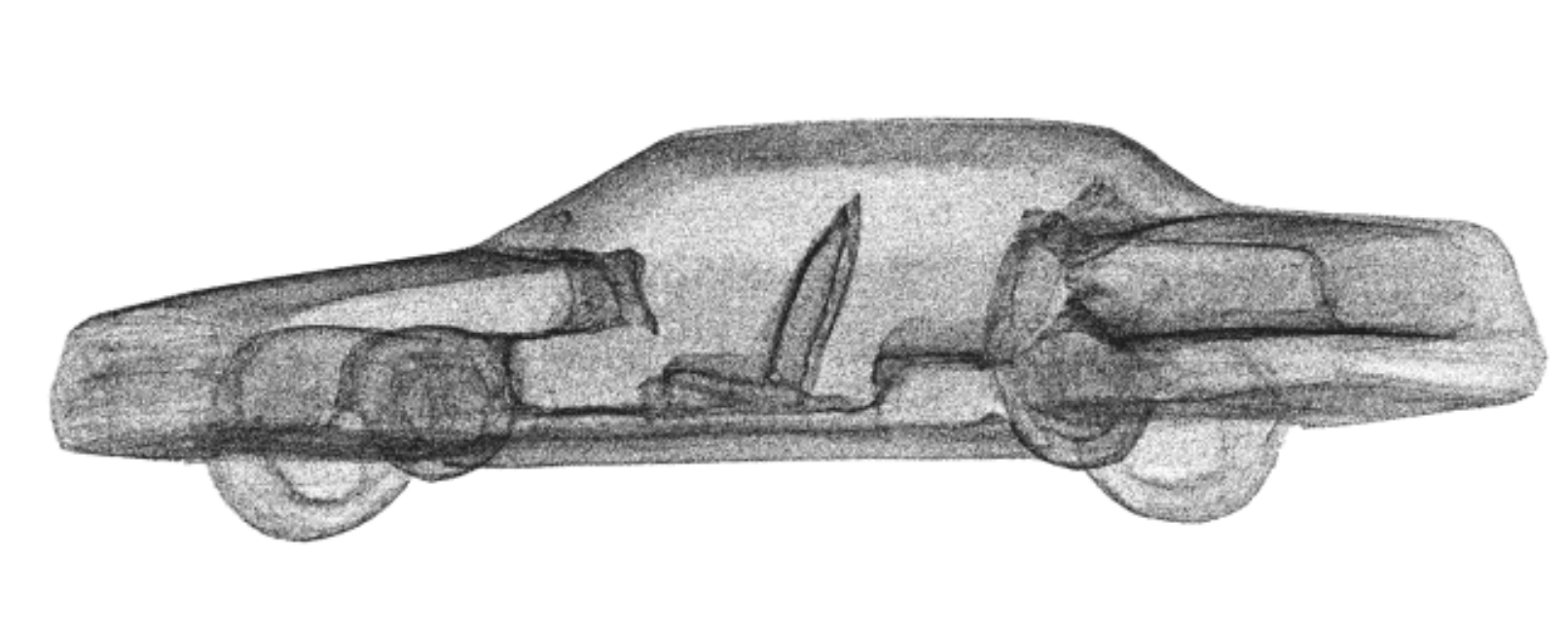}
      \includegraphics[trim={0.5cm 0.3cm 0.5cm 0.5cm},clip,width=0.1245\textwidth]{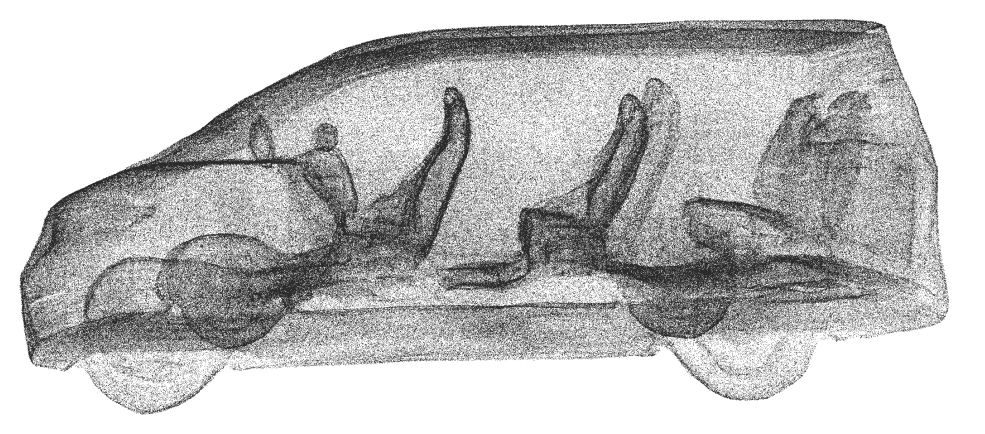}
    
    \includegraphics[trim={0.3cm 1cm 0.3cm 1cm},clip,width=0.1245\textwidth]{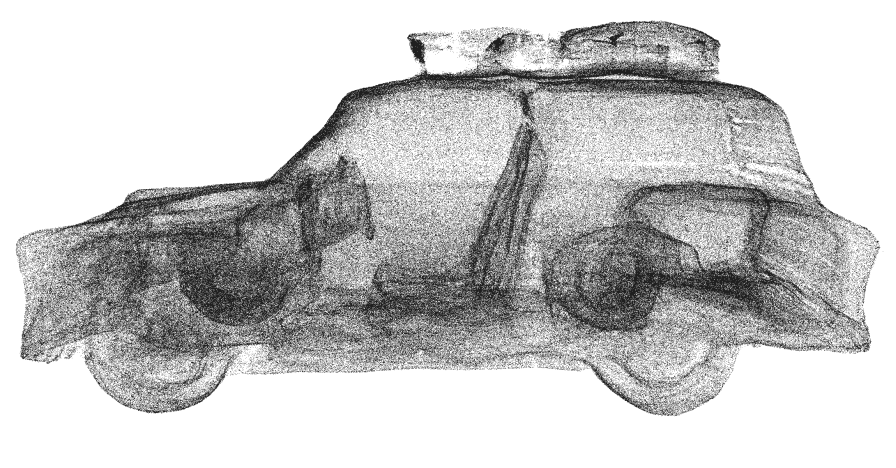}

     \includegraphics[trim={0.3cm 1cm 0.3cm 1cm},clip,width=0.1245\textwidth]{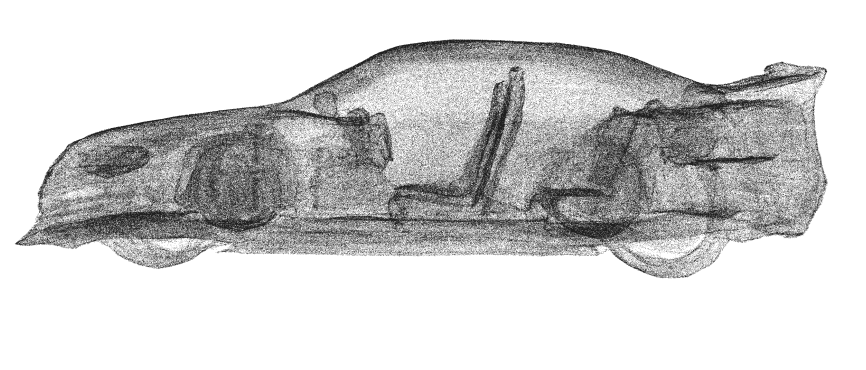}
    \\
     \includegraphics[trim={0.5cm 0.3cm 0.5cm 0.5cm},clip,width=0.1245\textwidth]{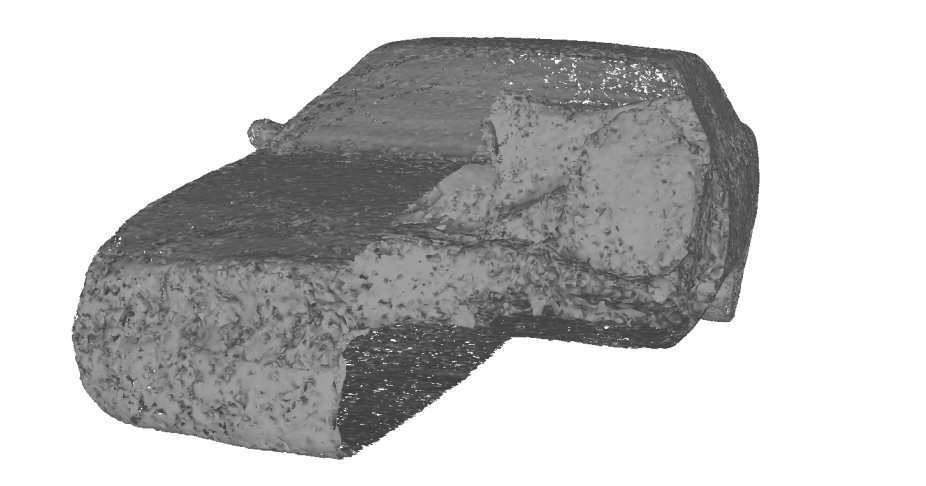}
       \includegraphics[trim={0.5cm 0.2cm 0.5cm 0.1cm},clip,width=0.1245\textwidth]{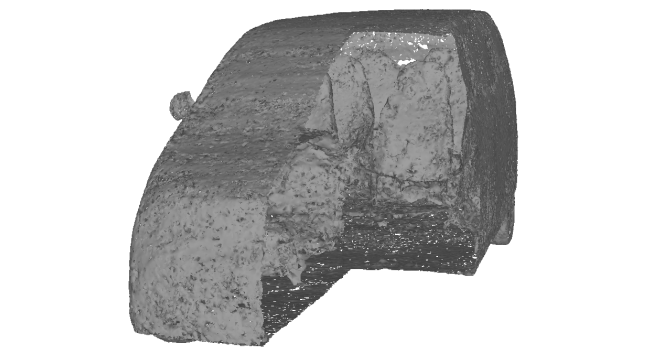}
        \includegraphics[trim={0.5cm 0.1cm 0.5cm 0.5cm},clip,width=0.1245\textwidth]{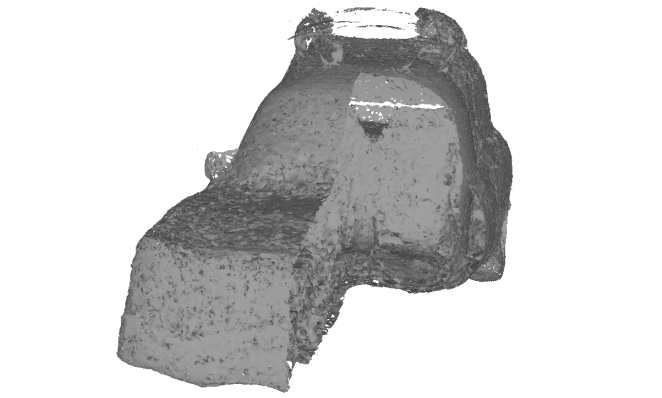}
        \includegraphics[trim={0.5cm 0.1cm 0.5cm 0.5cm},clip,width=0.1245\textwidth]{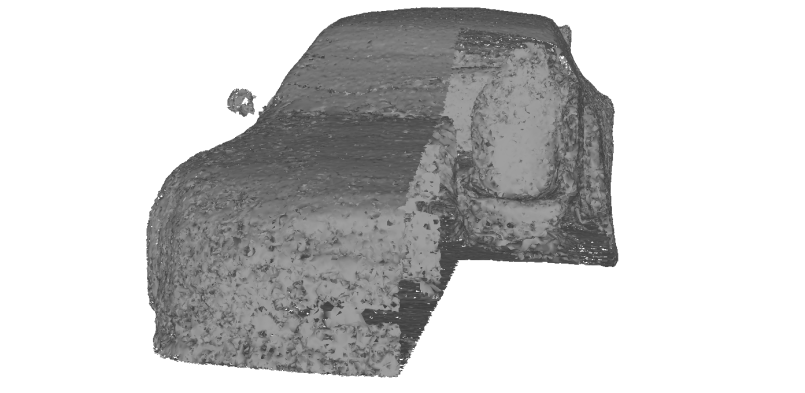}   
\end{tabular}}
\end{center}
\vspace{-1.5em}
\caption{\textbf{Generation:} Diverse generation results from our FullFormer model on the Full Cars dataset with internal structures. The high degree of detail of generated shapes is clearly visible in the dense point clouds. Note that, not only seats specific to car type, but also minute details such as steering wheels are well generated. High point clouds quality even allows to compute surface meshes (bottom) of the non-watertight shapes with internal structures.} 
\label{fig:internal}

\end{figure*}

\begin{figure}[ht!]
\begin{center}
\scalebox{0.87}{
    \vspace{-6.4em}

\begin{tabular}{ c  c  }
    \begin{subfigure}{\x\textwidth}
      \includegraphics[trim={0.0cm 0.0cm 0.0cm 0.0cm},clip, width=\textwidth]{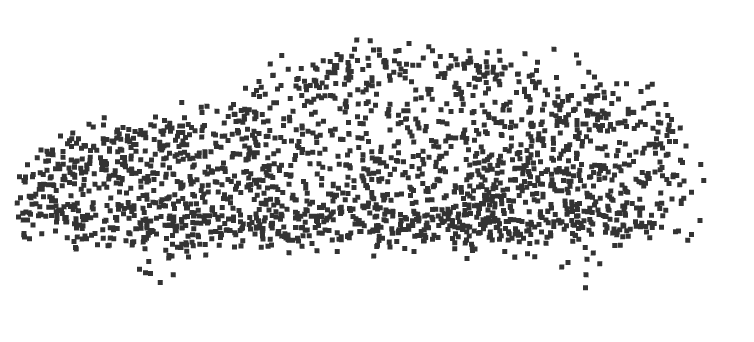}
    \end{subfigure}
     &  \begin{subfigure}{\x\textwidth}
      \includegraphics[trim={0.0cm 0.0cm 0.0cm 0.0cm},clip, width=\textwidth]{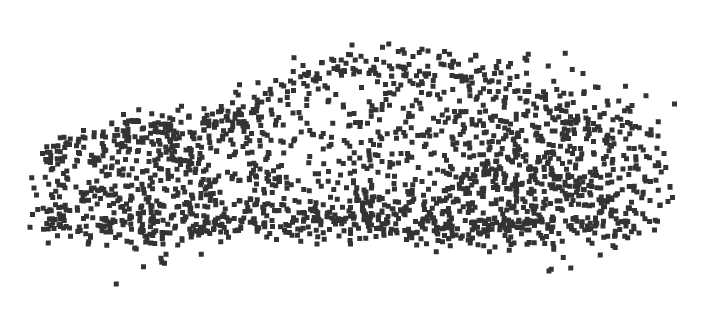}
    \end{subfigure}
    \vspace{-0.5em}
    \\
        
        \multicolumn{2}{c}{Diffusion~\cite{luo2021diffusion}} 
         \\
    \hline

    \\

        \begin{subfigure}{\x\textwidth}
      \includegraphics[ width=\textwidth]{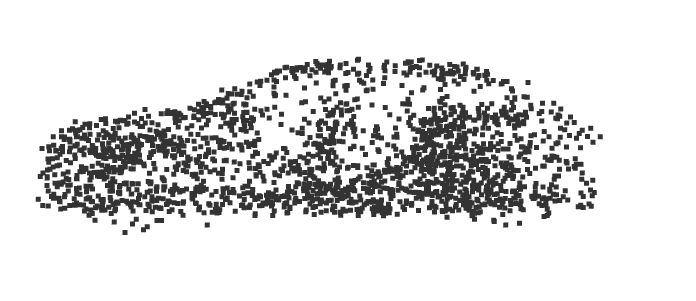}
    \end{subfigure}
    
    &
    \begin{subfigure}{\x\textwidth}
      \includegraphics[ width=\textwidth]{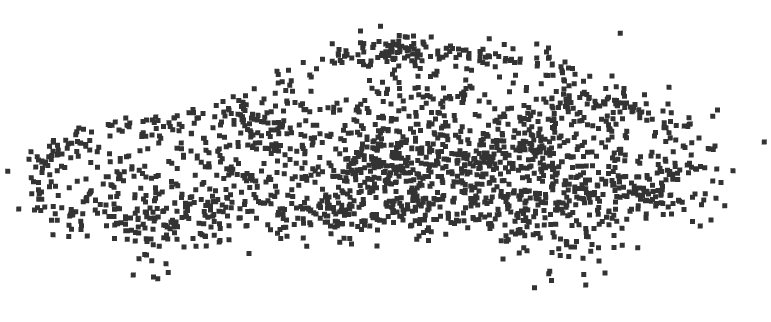}
    \end{subfigure}
       \vspace{-0.5em}
        \\
    \multicolumn{2}{c}{Point Flow~\cite{yang2019pointflow}} 
        \\
    \hline
    \begin{subfigure}{\x\textwidth}
      \includegraphics[trim={0.0cm 0.0cm 0.0cm 0.0cm},clip, width=1.0\textwidth]{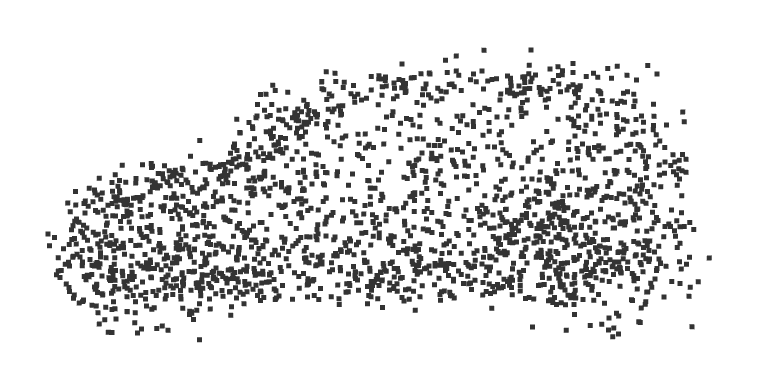}
    \end{subfigure}
     &  \begin{subfigure}{\x\textwidth}
      \includegraphics[trim={0.0cm 0.0cm 0.0cm 0.0cm},clip, width=1.0\textwidth]{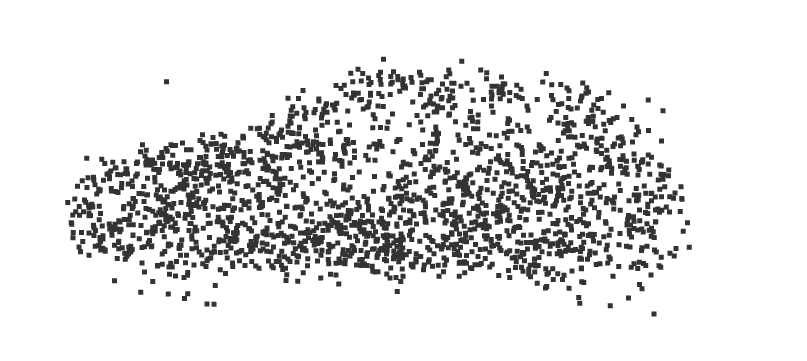}
    \end{subfigure} 
        \vspace{-0.5em}
    \\
    \multicolumn{2}{c}{Graph-CNN GAN~\cite{valsesia2019learning}} 
        \\
    \hline
    \begin{subfigure}{\x\textwidth}
      \includegraphics[trim={0.2cm 0.2cm 0.2cm 0.2cm},clip, width=1.1\textwidth]{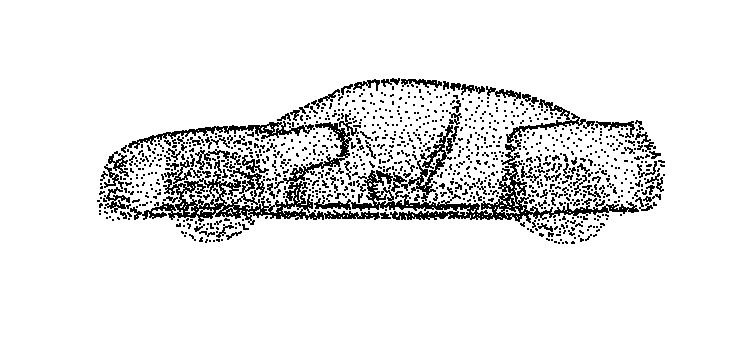}
    \end{subfigure}
     &  \begin{subfigure}{\x\textwidth}
      \includegraphics[trim={0.2cm 0.2cm 0.2cm 0.2cm},clip, width=1.0\textwidth]{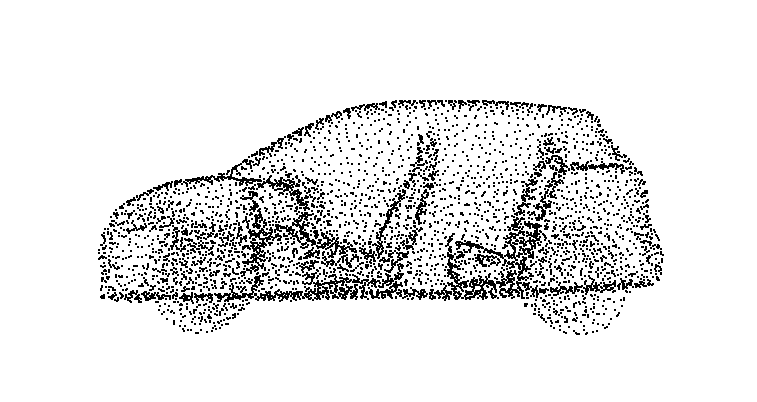}
    \end{subfigure} 
       \vspace{-0.5em}
    \\
    \multicolumn{2}{c}{FullFormer (Ours)}
    
            \vspace{-1.5em}

\end{tabular}}
\end{center}
\caption{\textbf{Generation Comparison:} Our model (with $16^3$ latent space resolution) shows high-quality internal structure generation results compared to previous models. It is apparent that these models do not achieve discernable internal structure. All point clouds are sampled to 2048 points.}
\label{fig:inner_h}
        \vspace{-1.5em}

\end{figure}

\subsection{Qualitative Results}
In this section, we show the qualitative performance of our generative model on the considered datasets.
\paragraph{ShapeNet}
The samples of point cloud generation results with 2048 points of our model against baseline models for the classes \textit{chairs} and \textit{airplanes} are presented in Fig.~\ref{fig:outer}. We highlight that our model does not rely on any priors in the form of preset tokens in the input sequence, thus ensuring the complete unconditioned generation of the results. The performance of our method is apparent with less noisy and realistic shape generations. We further note that immense diversity is present in the shapes generated, whereby all generated samples in Fig.~\ref{fig:outer} are of distinct visual designs. High fidelity is also perceptible across the generated examples. More results of generated samples of \textit{cars} are provided in the supplementary material.

\paragraph{Full Cars} 
We use the Full Cars dataset to showcase the veracity of our approach's key feature to generate high-fidelity outer shells with intricate internal geometric details. The qualitative results of randomly generated cars are presented in Fig.~\ref{fig:internal} demonstrating the efficacy of  our model in generating samples with rich internal geometric structures. Additionally, generated cars in Fig.~\ref{fig:internal} demonstrate a remarkable  level of diversity, for example, varied genres of cars with different number of seats. Fig. \ref{fig:inner_h} presents comparative results of randomly generated cars from Diffusion \cite{luo2021diffusion}, PointFlow \cite{yang2019pointflow} and our FullFormer. Both comparative methods are inherently capable of encapsulating internal structures, therefore are directly comparable. We retrain both methods on the `Full Cars' dataset. Our approach achieves a clear visual superiority over comparative methods, which fail to generate any discernible internal structure. It is also important to note that shapes in the training data lack dense internal geometries of high fidelity. Despite this limitation, our method is able to learn a general model which is capable of generating shapes with internal structures given noisy real-world raw data.


  
\subsection{Quantitative Results}
In this section, we present a quantitative evaluation of our model's performance in point cloud generation. The metrics discussed in section \ref{metrics} are tabulated in Table \ref{tab:quant}. Our method achieves state-of-the-art performance on all the metrics for the `Full Cars' dataset, validating the capability of FullFormer in generating complete shapes with rich insides. High coverage and low JSD further demonstrate that generated models exhibit high diversity which we also observe visually.

Moreover, we achieve the best performance in MMD and coverage across all classes of cars, chairs, and planes of the ShapeNet dataset compared with other baselines. While it is true that FullFormer appears to achieve higher JSD values than PointFlow \cite{yang2019pointflow} and Diffusion \cite{luo2021diffusion} for the ShapeNet dataset,  qualitative results continue to show diversity in all the considered datasets. Therefore the lower score of JSD for the ShapeNet dataset is hypothesized to be a cause of reference set selection.


\subsection{Limitations}
Unlike the high-fidelity achieved on outer shells, generated internal details exhibit lower quality. A sampling of the feature space limits the details of the shape's geometry. However, our approach presents the first effort towards generating internal details, which can be clearly seen in the presented qualitative results. Our evaluation is also constrained by the scarcity of available shape datasets with rich internal structures. Furthermore, we used off-the-rack methods to mesh our dense point cloud results which degraded the quality of our results, as there is no direct algorithm to mesh 3D shapes from unsigned distance fields. Especially on fine details and thin structures, the quality of generated shapes is not easy to assess from point clouds.

\section{Conclusion}
In this work, we present FullFormer: a model to generate 3D objects with internal structures. Our approach employs a vector quantized autoencoder (VQUDF) to learn 3D shape geometry. The encoder consumes a voxelized point cloud as input whilst the decoder predicts an unsigned distance field (UDF) of the 3D shape. To generate discrete embeddings of the 3D shape, we employ a latent transformer model. This transformer is trained autoregressively on indices of quantized shape embeddings learned by the VQUDF, making it computationally efficient. The trained transformer is then able to generate latent codes unconditionally. Generated codes are decoded into a UDF as the output representation ensuring that generated shapes have rich internal structure and high-fidelity outer surface at arbitrary resolution. We demonstrate superior qualitative and quantitative results compared to previous state-of-the-art methods. 


\clearpage
 
{\small
\bibliographystyle{ieee_fullname}
\bibliography{egbib}
}

\end{document}